\documentclass{article}
\usepackage{PRIMEarxiv}

\usepackage[utf8]{inputenc} 
\usepackage[T1]{fontenc}    
\usepackage{hyperref}       
\usepackage{url}   
\usepackage[normalem]{ulem}
\usepackage{booktabs}       
\usepackage{amsfonts}       
\usepackage{nicefrac}       
\usepackage{microtype}      
\usepackage{lipsum}		
\usepackage{graphicx}
\usepackage{amsmath}
\usepackage{doi}
\usepackage{comment}
\usepackage{caption}
\usepackage{cite}
\usepackage{subcaption}
\usepackage{amssymb}
\usepackage{makecell} 
\usepackage{authblk}
\usepackage[english]{babel}
\usepackage{csquotes}
\usepackage{xcolor} 
\usepackage{tikz}

\usepackage{subcaption}
\usepackage{xcolor}
\usepackage{multirow}

\usepackage{pgfplots}
\usepackage{pgfplotstable}
\pgfplotsset{compat=1.17}
\usepackage{pgfplots}
\usepgfplotslibrary{colorbrewer}
\pgfplotsset{compat = 1.15, cycle list/Set1-8} 
\usetikzlibrary{pgfplots.statistics, pgfplots.colorbrewer} 
\usepackage{pgfplotstable}
\usepackage{filecontents}

\newcommand{\bs}[1]{\boldsymbol{#1}}

\begin{document}

\twocolumn[
\begin{@twocolumnfalse}

\title{MeshGraphNet-Transformer: Scalable Mesh-based Learned Simulation for Solid Mechanics}

\author{
  M.\,M.~Iparraguirre,
  I.~Alfaro, D.~Gonz\'alez,
  E.~Cueto\\
  Keysight-UZ Chair of the National Strategy on Artificial Intelligence,\\
  Aragon Institute of Engineering Research (I3A),
  Universidad de Zaragoza, Zaragoza, Spain.\\
  \vspace{0.3cm}
  \texttt{\{mikel.martinez,iciar,gonzal,ecueto\}@unizar.es}
}


\maketitle
\begin{abstract}
\vspace{0.5cm}  

We present MeshGraphNet-Transformer (\mbox{MGN-T}), a novel architecture that combines the global modelling capabilities of Transformers with the geometric inductive bias of MeshGraphNets to preserve mesh-based graph representations. \mbox{MGN-T} overcomes the key limitation of standard Message Passing Neural Networks (MPNNs), known as \textit{under-reaching}, particularly in solid mechanics of high-resolution mesh simulations, where long-range information fails to be propagated along the mesh.

By directly capturing long-range physical interactions, \mbox{MGN-T} eliminates the need for deep message-passing stacks, without reducing resolution, as no coarsening or hierarchical representation is required. Instead, a physics-attention Transformer serves as a global processor, updating all nodal states simultaneously. The proposed feature engineering enables the successful simulation of impact dynamics, which, thanks to the mesh-based graph representation, naturally handles varying geometries, boundary conditions, contact and self-contact.

As part of the work, we further provide insights into the learning process by analysing and interpreting the physical tokens $P$. Additionally, we conduct ablation studies and extrapolation experiments to assess the robustness and physical consistency of the simulated solutions.

We demonstrate the scalability of the method, making it the first graph-based architecture to simulate industrial-level problems with realistic multi-material behaviour up to 100~k-node meshes. \mbox{MGN-T} reduces simulation time by a $10^3$ factor when compared to FEM, reporting relative errors way below 5\%. Compared with other baselines on the Deforming Plate benchmark, our method ranks second-best in accuracy, with a remarkable four-times reduction in trainable parameters.

\end{abstract}

\vspace{0.75cm} 

\end{@twocolumnfalse}
]

\section{Introduction}

Traditional solvers in solid mechanics are predominantly based on the finite element method (FEM), which can provide high-fidelity solutions when coupled with accurate physics-based models, but at a high computational cost, especially for many-query problems. Their accuracy depends critically on fine spatial meshes and, in explicit ./tdynamics, small time steps: insufficient spatial resolution may cause an undesirable level of error, while coarse temporal discretisation can lead to solutions that deviate from the governing partial differential equations, causing non-compliance with energy conservation laws for purely numerical reasons. Achieving reliable accuracy, therefore, requires very fine discretisation, significantly increasing computational demands.  Knowledge is embedded in the physical models, which describe point-level behaviour through constitutive equations, and in the PDEs and boundary conditions defined by the engineer. Deep learning takes this one step further by enabling models to learn from simulation results, effectively capturing and reusing prior knowledge to accelerate future predictions by simulating at lower temporal resolutions by several orders of magnitude. While this approach can substantially speed up simulations, it does not yet fully replace FEM, as it still requires high-fidelity data for training and, in general, generalises reliably only to scenarios similar to those seen during training \cite{RAISSI2019686,hernandez2021structure,tomasetto2025reduced}.  

\begin{figure}[t]
    \centering
\includegraphics[width=\columnwidth]{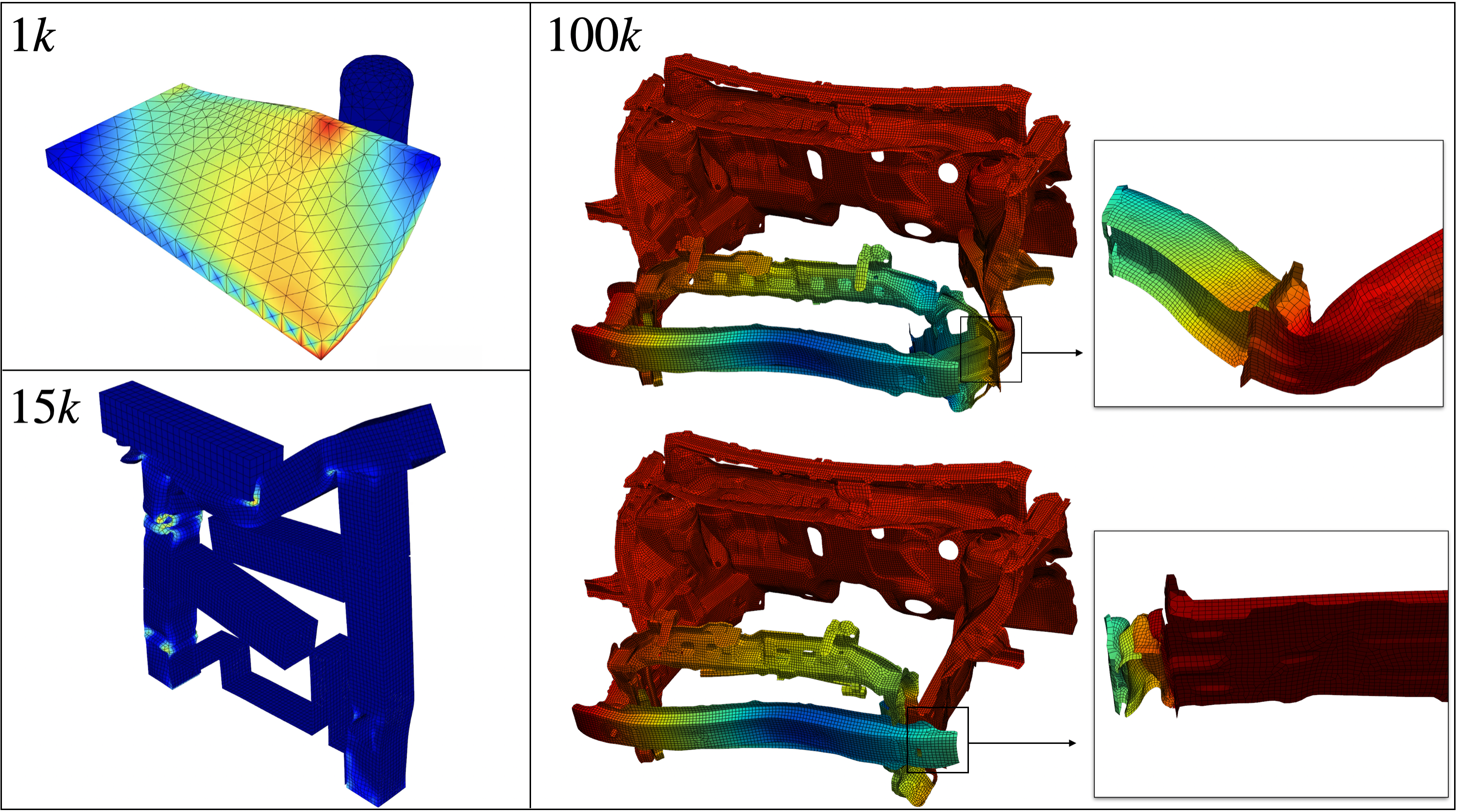}
    \caption{MeshGraphNet-Transformer predictions of test samples for the Deforming Plate, Pi-Beam and BIW crash benchmarks. Colour maps: von Mises $\sigma_{\text{VM}}$, plastic hardening $\alpha$ and displacements $\bs{u}$, respectively.}
    \label{fig:tisser}
\end{figure}

\begin{figure*}[h!]
    \centering
    \includegraphics[width=\linewidth]{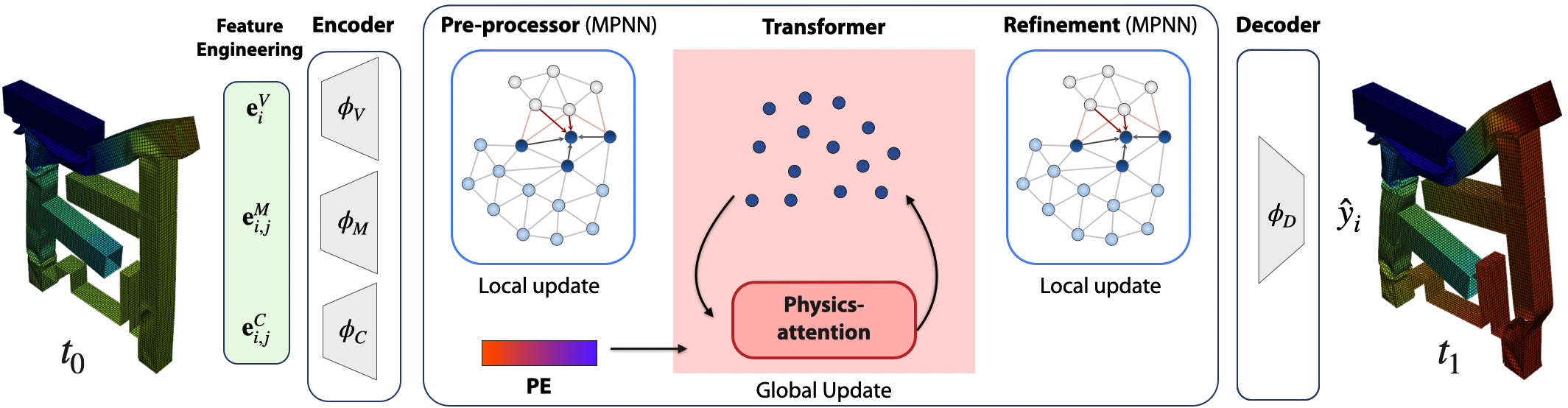}
    \caption{Illustration of the MGN-T architecture. Processor: Pre-processor MPNN (local update), Transformer processor with physics-attention (global update) and Refinement processor MPNN (local update).}
    \label{fig:mgnt}
\end{figure*}

This work focuses on learning from high-fidelity simulations by leveraging physics knowledge during training, testing the limits of purely data-driven methods, and exploring their ability to learn physical behaviour without explicit constraints. Graph Neural Networks (GNNs) naturally operate on graph-structured inputs, as models obtained using finite elements already have this graph structure. Sanchez-Gonzalez~\textit{et al.}~\cite{AGonzalezGNS} first applied GNNs to physics simulation in a particle-based setting, while Pfaff~\textit{et al.}~\cite{Pfaff2020MeshGraphNets} extended this to Lagrangian meshes with MeshGraphNets (MGN) for quasi-static collisions. Other examples of similar architectures include \cite{franco2023deep,roznowicz2024large,gorpinich2025bridging,di2025physics,pichi2024graph}. This mesh-based architecture can explicitly model dynamic interactions such as collisions and self-contact, introducing a strong local bias in its design and making it well-suited for simulating solid mechanics problems.

These architectures rely on Message Passing Neural Networks (MPNNs), which compute predictions through iterative node-to-node exchanges rather than processing the full graph globally \cite{brandstetter2022message}. Each node updates its state by aggregating information from its neighbours, providing a strong local bias that mimics the local character of the equations involved (e.g, Elasticity). However, this locality is also the main limitation: physical phenomena must be within the effective message-passing reach, making the number of message-passing iterations a critical hyperparameter for the model’s performance; otherwise, it will suffer from under-reaching, as shown by Tesan~\textit{et al.}~\cite{under-reach-mpi}. This node-centric formulation provides strong geometric flexibility, allowing MGN models to generalise to unseen meshes given node features and connectivity. However, long-range interactions are captured only when a sufficient number of message-passing layers is used, which becomes intractable for large meshes \cite{tierz2025feasibility}.

To overcome this limitation, we present MeshGraphNet-Transformer, see Fig. \ref{fig:tisser}, a novel architecture that combines the global modelling capabilities of Transformers with the geometric inductive bias of MeshGraphNets to simulate solid mechanics problems. 

The main contributions of this work are:
\begin{itemize}
\item \textbf{MeshGraphNet-Transformer (MGN-T):} An extension of the MeshGraphNet architecture that removes the message-passing under-reach bottleneck, see Fig.~\ref{fig:mgnt}. The pre-processor and refinement MPNN blocks enable local interactions, while a Transformer physics-attention processor serves as a global update.

\item \textbf{Preserving mesh-based structure:} It keeps the powerful mesh-based representations and the feature engineering of MGN, to explicitly model collisions and self-contact.
\item \textbf{Scalability and generalisation :} MGN-T generalises to different geometries, mesh topologies, and boundary conditions, simulating highly non-linear phenomena such as plasticity. The method scales from academic benchmarks (1~k-node problems) to industrial-scale problems (15~k- and 100~k-node meshes). 
\end{itemize}

\section{Background}

Several studies have attempted to improve the performance limitations of standard GNNs. Variants that incorporate thermodynamic inductive biases and improve predictive accuracy are presented in~\cite{hernandez2022thermodynamics,tierz2025graph,zhang2022gfinns,gruber2023reversible,hernandez2023port,bermejo2025meshgraphnets}.

To overcome under-reaching, BSMS-GNN~\cite{BSMS-GNN-cao23a} constructs a hierarchical graph by bi-stride pooling, which groups nodes every other topological depth, connected via downsampling and upsampling layers. The coarse levels are computed prior to training and are non-adaptive. However, while effective for Directed Acyclic Graphs, it may fail to capture suitable hierarchical structures in complex meshes, limiting its applicability to industrial-scale simulations. Similarly, ROBIN, proposed by Würth et al.~\cite{robin2025}, combines diffusion processes with AMB-based pooling to preserve high-frequency information and node density. Despite achieving strong performance, it is still limited to 2D benchmark problems. EvoMesh~\cite{EvoMesh-deng25} introduces a differentiable multi-scale graph construction in which a learnable probability determines whether nodes are retained or discarded at coarser levels. Anisotropic message-passing layers are used to facilitate communication across scales. This approach enables an adaptive, learnable hierarchy that overcomes key limitations of fixed coarsening strategies, yet it has not been tested on industrial-scale problems.

More recently, M4GN~\cite{M4GN_LeiBo2025} is presented as a hierarchical graph surrogate that operates on meshes segmented via a hybrid mesh–graph scheme. It requires an explicit preprocessing step to compute the mesh–graph segmentation prior to training.  They employ a 2-step hybrid segmentation strategy that combines graph partitioning with a superpixel-style refinement step, guided by modal-decomposition features to ensure physical consistency. An advantage of our approach is that it is end-to-end trainable, requiring no preprocessing or segmentation. When tested on several benchmarks, their application remains far from industry-scale problems in solid mechanics, where larger meshes ($N >>$ 1~k) are used.

For more industry-like problems, ReGUNet, introduced by Li~\textit{et al.}~\cite{Li2025ReGUNet}, extends MeshGraphNets with a recurrent hierarchical U-Net and was the first to demonstrate crashworthiness simulation on an automotive B-pillar. While its hierarchical design and history handling mitigate error accumulation, the reliance on a fixed coarse mesh limits its ability to adapt to datasets where no coarser meshes are available. Overall, hierarchical approaches partially alleviate the limitations of deep message passing. However, they require coarser representations, which are often difficult to automate when high-detail geometries must be simulated. Although message-passing under-reaching is diminished, it is not eliminated, as you will still need to propagate information at the coarsest level. As an advantage over these methods, ours is hierarchy- and coarsening-free, reducing preprocessing steps.

As an alternative to MPNN-based methods, Transformers have proven effective for handling large numbers of nodes, since they process nodes in parallel and can capture global phenomena without iterative message passing. Y. Yu~\textit{et al.}~\cite{HTCM2024} introduced HCMT, a mesh-based method that employs a learnable pooling strategy with transformers. 

Two works are particularly relevant to ours. Transolver~\cite{HaixuWu2024} introduced physics-attention and slicing mechanics, allowing a tokenisation that overcomes the limitation of fixed node counts. Its extension, Transolver++~\cite{HuakunLou2025}, further improves physics-attention through a learnable eidetic slicing technique, enabling sharper weights. Despite their strong capability to handle long-range interactions on large meshes, most of the existing Transformer-based approaches are meshless, operating only at the node level, ignoring important structural information such as mesh connectivity, self-contact and collisions, which our method MGN-T preserves. Universal Physics Transformer~\cite{ABUPT2025} is a Transformer-based neural operator for volumetric CFD meshes that captures boundary and interior structure via anchor-attended tokens, but has not been applied, to the best of our knowledge, to solid mechanics problems.

The outline of the paper is as follows. In Section \ref{method}, we present the method. In Section \ref{sec:datasets}, we introduce the numerical examples used in Section~\ref{sec:experiments}, in the experimental phase. As a closure of the research, a discussion in Section~\ref{discussion}.

\section{Methodology}\label{method}

We focus on developing a neural PDE solver for solid mechanics trained on synthetic data coming from high-fidelity simulations. The solver must handle high-resolution meshes in which the number of nodes, geometries, and boundary conditions may vary and, additionally, collision interactions and self-contact on highly non-linear materials, including elasto-plastic behaviour. 

Temporal evolution is simulated using an autoregressive approach that recursively integrates 1-step model predictions.

\subsection{Feature engineering}
The physical system is represented as a multi-graph $G(V, E_M, E_C)$, where each structure corresponds to a subgraph (i.e., different parts of the model). Nodes correspond to mesh vertices $V$, and mesh edges $E_M$ correspond to mesh connectivity among components, including tied ones. The subgraphs are dynamically connected through contact edges $E_C$ when contact occurs. Tied constraints are computed detecting $ k$-nearest-neighbours over the assembled components.

Physical variables differ across problems, emphasising the importance of selecting the appropriate nodal variables to encode the initial and boundary conditions, as these will determine the solution of my problem.

\textbf{Node features:}
Each node $i$ has a state vector $\bs{s}_i$ containing problem-specific physical variables or parameters (e.g., velocity or shell thickness), and a one-hot encoded vector which indicates the node type $\bs{n}_i$ (e.g., deformable component or rigid body). The node feature vector is defined as $\bs{e}^V_i = (\bs{s}_i, \bs{n}_i)$.

\textbf{Edge features:}
We encode geometric information using relative positional variables to maintain spatial equivariance. We consider two types of edges:
\begin{itemize}
\item Mesh edges: $\bs e^{M}_{ij} \in E_M$ defined by mesh connectivity and tied constraints. For each edge $(i,j)$, we encode relative distances between connected nodes in the reference $\bs{X}$, and current configurations $\bs{x}$:
\[
\bs{e}^{M}_{ij} =
(\bs{X}_i - \bs{X}_j,\,
\|\bs{X}_i - \bs{X}_j\|,\,
\bs{x}_i^{t} - \bs{x}_j^{t},\,
\|\bs{x}_i^{t} - \bs{x}_j^{t}\|)
\]

\item Contact edges: $\bs e^{C}_{ij} \in E_C$ defined by nodes within a contact radius $r_c$, if $\|\bs e^{C}_{ij}\| < r_c$. This includes both collision contact and self-contact. For each contact edge, we encode the relative distance between connected nodes in the current configuration:
\[
\bs{e}^{C}_{ij} =
(\bs{x}_i^{t} - \bs{x}_j^{t},\,
\|\bs{x}_i^{t} - \bs{x}_j^{t}\|).
\]
\end{itemize}

\textbf{Positional encoding:}
We construct a stationary-wave representation over the component’s spatial domain in the undeformed configuration. This provides positional information about the simulated structure during the global update, with obstacles---rigid wall and actuator---being ignored. Sinusoidal functions are defined along the length, width, and height of the component \cite{attention_Vaswani2017}, making it translation-invariant.

\subsection{Architecture}
\label{sec:hierarchical}

The proposed MeshGraphNet-Transformer (MGN-T) architecture follows the Encoder–Processor–Decoder scheme illustrated in Fig.~\ref{fig:mgnt}.  Unlike standard MeshGraphNets, in which the Processor consists of a deep stack of message-passing layers, MGN-T replaces this iterative structure with a three-block processor: a \textit{Pre-processor}, a \textit{Transformer processor}, and a \textit{Refinement processor}.

\textbf{Encoders}: Three dedicated encoders map the node and edge {features} into a high-dimensional latent space :
\[
\bs{\varepsilon}^{V}_{i} = \phi_V(\bs{e}^V_{i}), \quad
\bs{\varepsilon}^{M}_{ij} = \phi_M(\bs{e}^{M}_{ij}), \quad
\bs{\varepsilon}^{C}_{ij} = \phi_C(\bs{e}^{C}_{ij}),
\] where each $\phi$ is a 2-layer MLP with LeakyReLU activations, and layer normalisation.

\textbf{Pre-processor}: implemented as MPNN. This block is responsible for absorbing boundary conditions and local interactions. Gathers information from connected nodes, using information from the mesh and contact edges, before the global update. 


\textbf{Transformer processor}: Performs a global update of node features using stacked Transformer blocks with physics attention. Before being fed into the Transformer, node attributes are concatenated with positional encodings and projected to match the multi-head attention. The slicing mechanism of the physics-attention projects the $N$ nodes into $P$ latent physical tokens. By operating on these tokens, the Transformer enables efficient information exchange across them. Then, the de-slicing mechanism projects $P$ back onto $N$ mesh nodes, efficiently completing the global update of the mesh. Highlight that only simulated structure nodes are considered here, not obstacles.

\textbf{Refinement processor}: it reinforces geometric inductive biases and local consistency prior to decoding. It is also implemented as MPNN.

\textbf{Decoder}: Projects the updated latent node features $\bs{\varepsilon}^{V}_{i}$ to the physical space, estimating the variables of interest.
$\bs{\hat{y}}_{i} =\phi_D(\bs{\varepsilon}^{V}_{i})$.

\subsection{Physics-attention with eidetic states}
\label{sec:attention}
We adopt the physics-attention mechanism introduced in Transolver~\cite{HaixuWu2024} and its extension with eidetic states proposed in Transolver++~\cite{HuakunLou2025}. Their proposed physics-attention reduces the quadratic complexity $\mathcal O(N^2)$ of standard self-attention by projecting $N$ node features into $P$ physical tokens, as $\bs{z}$ eidetic states. This projection not only reduces computational cost but also allows easy handling of varying input sizes and efficient scaling to a large mesh, as the complexity now is $\mathcal O(P^2)$, with $P \ll N$. To this, we must add the overhead of having the pre-processor and refinement MPNN blocks $\mathcal O(L(N+E))$, where $L$ is the total MPI and $E$ edge connections.

In our approach, rather than encoding the entire system $N$ into the physics-attention mechanism, we restrict the representation to the simulated structure only. Interactions with obstacles, such as actuators or rigid walls, are absorbed during the preprocessing stage through contact edges. This design encourages the model to learn collision dynamics explicitly, improving generalisation to varying boundary conditions. This allows the physics-attention mechanism to focus solely on the dynamics of interest.

To generate the projection, first, a local adaptive temperature mechanism dynamically adjusts the sharpness of the slice weight $\bs{w}$ between each node and the corresponding physical token. The temperature $\tau$ is learnt from the node features $\bs{x}$, with $\tau = \{\tau_i\}_{i=1}^{N} = \{\tau_0 + \text{Linear}(x_i)\}_{i=1}^{N}$, to increase expressivity to learn relationships between nodes and physical tokens, 
\begin{equation*}
\{ \bs{z}_j \}_{j=1}^{P} 
= 
\left\{
\frac{\sum_{i=1}^{N} w_{ij} \bs{x}_i}{\sum_{i=1}^{N} w_{ij}}
\right\}_{j=1}^{P},
\end{equation*}
where
\begin{equation*}
\bs{w} 
= \mathrm{Softmax}\!\Bigg(
\frac{\mathrm{Linear}(\bs{x}) - \log(-\log(\epsilon))}{\tau}
\Bigg).
\end{equation*}

Second, the slice reparameterization mechanism with the Gumbel–Softmax trick~\cite{Jang2017} is used to calculate the weights $\bs{w}$. This enables differentiable sampling from the categorical distribution over physical states $\bs{z}$, where the Gumble noise is a function of $\epsilon \sim \mathcal{N}(0, 1)$. These two techniques, proposed in~\cite{HuakunLou2025}, enable sharper weighting and prevent tokens from collapsing, providing a better latent representation of the physical token states, with $\tau_0=1.0$ used in the literature.

Physics-attention is defined as a multi-head attention mechanism \cite{attention_Vaswani2017} over the $P$ physical tokens, where $P$ is a hyperparameter, tuned during model selection. Each token is linearly projected into Query ($Q$), Key ($K$), and Value ($V$), with dimensionality $c$. The scaled dot product between the ${Q}$ and the $K$ produces the attention weights, normalised with a $\mathrm{Softmax}$ function. Afterwards, the attention weights are projected onto the $V$ to generate the updated $\bs{z'}$ physical $P$ tokens,
\begin{equation*}
\begin{aligned}
Q, K, V &= \mathrm{Linear}(\bs{z}), \quad
\bs{z'}=\mathrm{Softmax}\!\left(\frac{QK^{T}}{\sqrt{c}}\right) V .
\end{aligned}
\end{equation*}

\subsection{Model training}
MGN-T is trained using teacher-forced 1-step predictions. Let $N$ be the number of nodes per trajectory and $B$ the batch size, representing snapshots from the same trajectory. The ground truth node outputs are denoted as $\bs{y}_i^b$ and the model predictions as $\hat{\bs{y}}_i^b$, where $i = 1,\dots, N$ and $b = 1, \dots, B$. It is important that batch samples belong to the same trajectory to ensure consistent matrix reshaping before applying the Transformer processor. The samples are shuffled to avoid any potential bias, and this only affects batch training. The training loss is defined as the mean squared error ($L_2$ loss), computed per node and averaged over the batch across deformable body nodes:
\begin{equation}
\mathcal{L} = \frac{1}{B} \sum_{b=1}^{B} \frac{1}{N} \sum_{i=1}^{N} \left\| \bs{y}_i^b - \hat{\bs{y}}_i^b \right\|_2^2.
\end{equation}

\section{Numerical Examples}

\label{sec:datasets}

We consider Lagrangian mesh-based models undergoing large deformations in two distinct regimes: a quasi-static scenario with the Deforming Plate benchmark (see Section~\ref{sec:deforming_plate_dataset}) and an impact dynamics scenario with the Pi-Beam and Body in White crash benchmarks (see Sections~\ref{sec:esi-dataset} and~\ref{sec:yaris_crash}, respectively).
These numerical examples cover 1~k-, 15~k-, and 100~k-node mesh simulations, proving the MGN-T capacity to deal with a wide range of sizes, reaching industrial-scale simulations (see Fig.~\ref{fig:tisser}).

\begin{table}[h]
\centering
\scriptsize
\setlength{\tabcolsep}{5pt}
\renewcommand{\arraystretch}{1.}

\begin{tabular}{lccccc}
\toprule
Dataset & Steps & Nodes & Mesh & $\bs{s}_i$ & $\bs{y}_i$ \\
\midrule

Def. Plate 
& 400 
& 1 k 
& Teth.
& $\bs{u}_i^{\text{pre}}$ 
& $\bs{u}_i^{k+1},\, \sigma_{\text{VM},i}^{k+1}$ \\

Pi-beam       
& 100 
& 15 k
& Rect.
& $\bs{v}_i^{t},\, \alpha_i^{t},\, k_i$ 
& $\bs{u}_i^{t+1},\, \bs{v}_i^{t+1},\, \alpha_i^{t+1}$ \\

BIW       
& 45 
& 100 k
& Rect.
& $\bs{v}_i^{t},\, \alpha_i^{t},\, k_i$ 
& $\bs{u}_i^{t+1},\, \bs{v}_i^{t+1},\, \alpha_i^{t+1}$ \\
\bottomrule
\end{tabular}

\caption{Dataset characteristics and target variables. Nodes are approximated values.}
\label{tab:datasets}
\end{table}

\begin{figure*}[h]
    \centering
    \includegraphics[width=\linewidth]{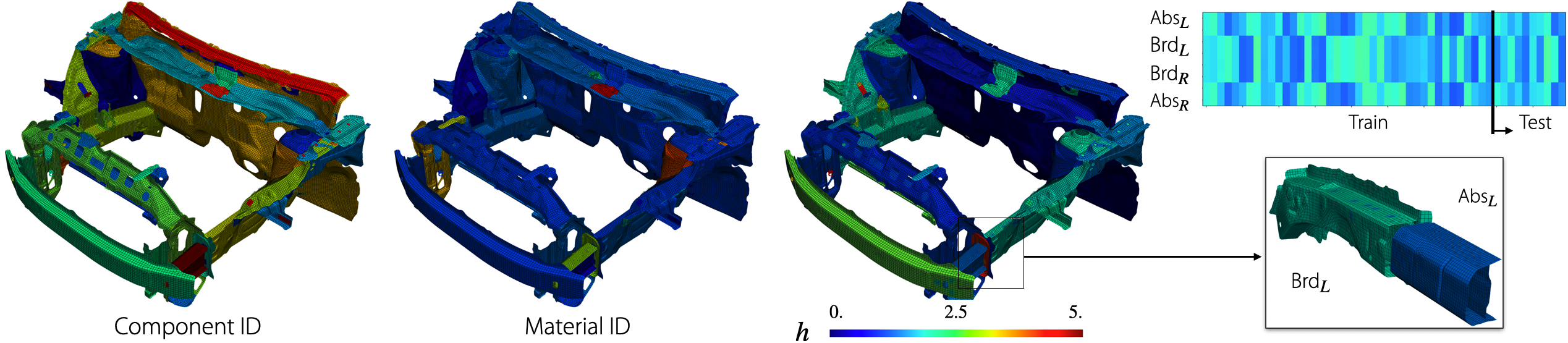}
   \caption{BIW crash example. Visualisation of component IDs, material IDs, and thickness $h$.  Highlighted the Absorber (Abs$_L$) and Bridge (Brd$_L$) parts, whose thickness varies across trajectories.  Diagram of different thickness setups; from the 40 onwards, test, previous train.}
    \label{fig:yaris-dataset}
\end{figure*}

\subsection{Deforming Plate benchmark}
\label{sec:deforming_plate_dataset}
This widely used public benchmark introduced by Pfaff \textit{et al.}~\cite{Pfaff2020MeshGraphNets} is designed to evaluate surrogate models in solid mechanics.
It consists of simulations of an actuator impacting a hyperelastic, deforming plate, in a quasi-static regime. The geometries are 3D discretised using tetrahedral elements, resulting in over 1000 nodes per simulation. The dataset aims to
investigate queries involving changes in the geometry, boundary conditions and mesh. Synthetic data were generated using COMSOL Multiphysics 6.0 (COMSOL, Inc.). 

\subsection{Pi-Beam {benchmark}}
\label{sec:esi-dataset}

\begin{figure}[h!]
    \centering
    \subfloat[Multi-component view]{%
        \includegraphics[width=0.45\linewidth]{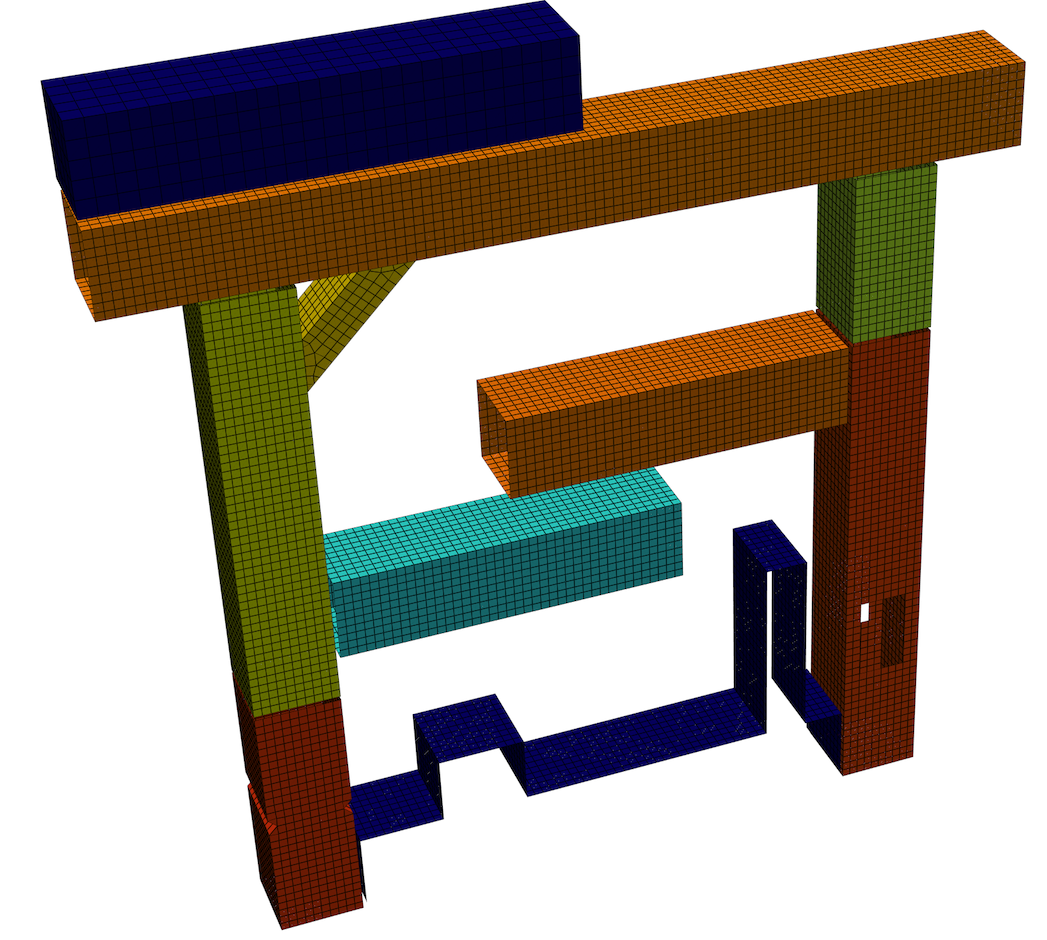}%
        \label{fig:p-beam_multicomponent}%
    }\hfill
    \subfloat[Mesh view]{%
        \includegraphics[width=0.45\linewidth]{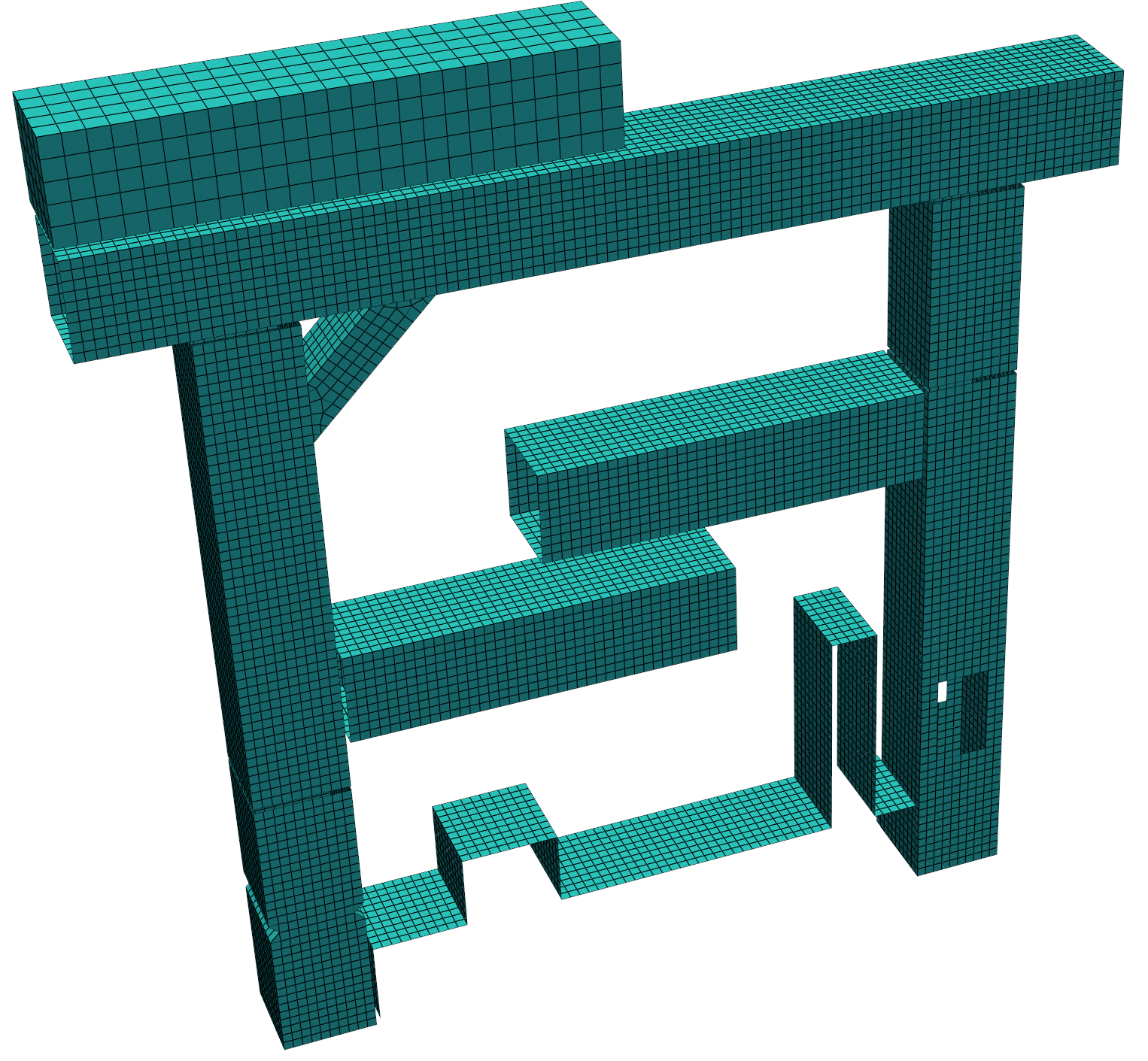}%
        \label{fig:p_beam_mesh}%
    }

    \caption{Pi-Beam example. 
    (a) Visualisation of the multi-component structure, where each colour represents a different component with a different thickness. 
    (b) Visualisation of the FEM mesh with rectangular shell elements.}
    \label{fig:p_beam_example}
\end{figure}

The Pi-Beam benchmark reproduces the impact of a deformable, elastic-plastic, multi-component energy absorber against a rigid obstacle, representing a realistic geometry automotive crash scenario. The learning task consists of predicting the full nonlinear dynamic response of the system, including contact, collisions, and subsequent spring-back, where each simulation corresponds to a varying component thickness setup \( h \in (0.1, 0.3) \).

Two materials are considered: one deformable material (steel,
Elastoplastic, with power-law hardening and Young’s modulus E = 210 GPa) and a rigid material. The multi-component structure consists of 9 tied components, where node connectivity is processed as a mesh connection, $\bs e^{M}_{ij} \in E_M$, each with a different thickness (see  Fig.~\ref{fig:p_beam_example}). 

The geometry is discretised using shell elements with a total of 16\,193 nodes per simulation (15\,641 deformable and 552 rigid obstacles). The simulation consists of 100 incremental steps subsampled at 1.5~ms per step. Node state variables $\bs{s}_i$ include the phenomenological internal hardening variable $\alpha$, which measures how much each node has plastified (see Table~\ref{tab:datasets}).

\subsection{Body in White crash benchmark}
\label{sec:yaris_crash}

This dataset reproduces the impact of a car body-in-white structure against a rigid wall, representing a real industrial-scale simulation in the automotive industry (see Fig.~\ref{fig:yaris-dataset}). The learning task consists of predicting the full nonlinear dynamic response of the system for given different thickness component setups in the absorber part $\{\text{Abs}_{L}, \text{Abs}_{R}\}$ and bridge $\{\text{Brd}_{L}, \text{Brd}_{R}\}$. The absorber is responsible for dissipating the energy of the collision, critical for safety in design. 

The simulation comprises 100 components and 17 different materials. The geometry is discretised using quadrilateral shell elements, with a total of 100~k nodes per simulation, of which 1200 belong to the rigid wall. As previously mentioned, the contiguous components are tied. During simulation, self-contact occurs, as well as impact against a rigid wall, both processed as contact edges, $\bs e^{C}_{ij} \in E_C$. The node type is stored as a one-hot encoder vector $\bs {n_i} $, encoding the 17 different material properties, from rubbers to different elastic-plastic materials.

Both Pi-Beam and Body in White (BIW) data were generated using the Pam Crash software (ESI Group, Keysight Technologies). Explicit finite element method (FEM) solver with a temporal resolution of $6.3 \times 10^{-4}$ ms, and mass scaling. The time cost per simulation is 10 and 50 minutes for the Pi-Beam and Body in White, respectively.  For the Pi-Beam example, we considered 28 different trajectories: 18 in train, 10 for test. For the Body in White crash example, we considered 50 trajectories: 40 for training and 10 for testing.

\section{Experiments}
\label{sec:experiments}
First, in Section~\ref{sec:exp_deforming}, we evaluate the performance of \mbox{MGN-T} against state-of-the-art approaches on the Deforming Plate benchmark problem in solid mechanics. We further investigate the physical interpretability of the method by analysing the learned physical tokens, providing insight into what they focus on within the domain.

Second, in Section~\ref{sec:exp_pibeam}, we assess the performance of \mbox{MGN-T} in the Pi-Beam problem, comparing it to MGN in scenarios where under-reaching is a key limitation. Additionally, we examine the physical consistency of the predicted solutions.

Finally, in Section~\ref{sec:exp_toyota}, we employ the Body in White crash dataset to conduct an ablation study evaluating the impact of the message-passing blocks, the positional encoder, and the $P$ tokens of \mbox{MGN-T}. Also, we analyse the model's extrapolation capabilities.

We evaluate the model's performance using the root mean squared error for 1-step predictions (RMSE-1) and the full-trajectory error (RMSE-all). Further details regarding error calculation are given in Appendix \ref{ap:errors}.

\subsection{Deforming plate} 
\label{sec:exp_deforming}
We explore the capability to simulate varying geometries, mesh topology, and boundary conditions using the Deforming Plate benchmark. For this experiment, the proposed architecture consisted of a processor with 2 MPNN blocks for preprocessing and refinement, each, and 2 transformer blocks with 16 physical tokens ($P=16$) and 4 heads on the multi-head attention. Trained during 25 epochs with batch size 32 and learning rate $10^{-4}$ with cosine annealing scheduler, reduced to $10^{-6}$.

\begin{figure}[t]
\begin{subfigure}{\linewidth}
    \centering
    \includegraphics[width=\linewidth]{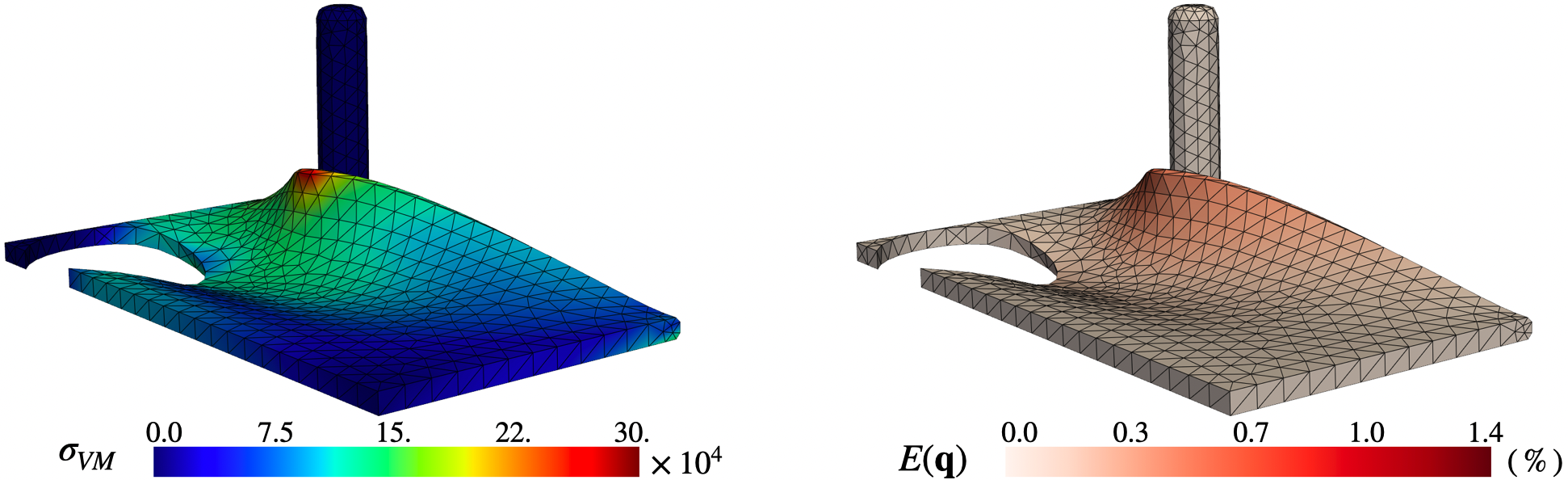}
    \caption{Test prediction and error. }
    \label{fig:prediction_plate}
\end{subfigure}

\vspace{0.5em}

\begin{subfigure}{\linewidth}
    \centering
    \includegraphics[width=\linewidth]{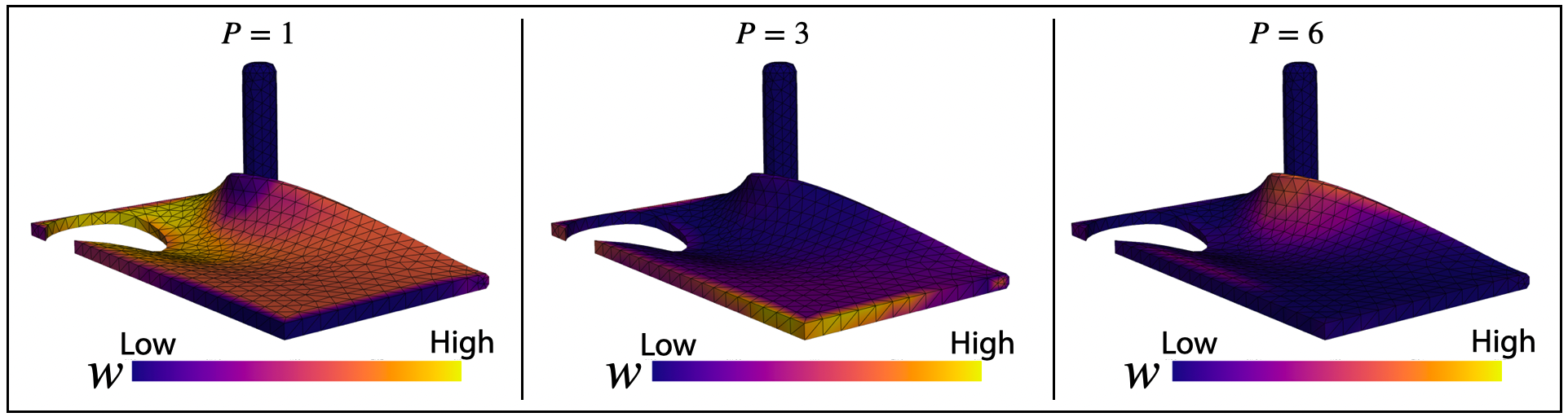}
    \caption{$P$ tokens}
    \label{fig:tokens_plate}
\end{subfigure}
    \caption{Prediction test sample \mbox{MGN-T} model. a) prediction $\sigma_{\text{VM}}$ and error distribution in positions $\bs{q}$. b) Token weights $\bs{w}$ visualisation.}
    \label{fig:deforming_tokens}
\end{figure}

We report a comparison table of state-of-the-art mesh-based baselines benchmarked on the dataset, allowing for a direct overview of the reported performance–complexity trade-offs across methods (see Table~\ref{tab:plate_combined}). M4GN and \mbox{MGN-T} are the best performers; both combine MPNNs with Transformers to handle long-range relationships and avoid under-reaching. In contrast to M4GN, our method can be trained end-to-end without the need for any hierarchical coarsening. In many cases, hierarchy or coarsening is highly domain- and geometry-dependent and is therefore nontrivial to implement. \mbox{MGN-T} achieves the second-best error while using four times fewer parameters. Notice that relative errors are below most industrial applications' requirements (see Table \ref{tab:deforming_plate_errors}). See Fig. \ref{fig:prediction_plate} for qualitative results.

\textbf{Physical Tokens $P$ interpretability}: While directly interpreting the learned token representations remains challenging, visualising the weights $\bs{w}$ provides qualitative insight into what the model has learnt under the hood (see Fig.~\ref{fig:tokens_plate}). The weights illustrate how information from the full mesh is aggregated into each $P$ token. From our perspective, these physics tokens are interpreted as a \emph{dynamic hierarchical representation} of the system.  This allows the model to construct an implicit hierarchy that evolves throughout the simulation, responding to changes in state variables, geometry and boundary conditions, rather than solely on geometry. 

As shown in Fig.~\ref{fig:deforming_tokens}, the model is capable of learning physical relationships across nodes and distinguishing them without overlap. $P=1$ weights the most deformed nodes, $P=3$ focuses on clamped nodes while $P=6$ highlights nodes that are in contact with the actuator (load).  Consistency in how tokens weigh across different geometries is observed (e.g. token $P=6$ always weights nodes in contact with the actuator). As the actuator's prescribed displacement conditions are absorbed by the preprocessor block, our method does not need to allocate tokens $P$; the physics-attention focuses only on deformable nodes.

\begin{table}[t]
\centering
\scriptsize
\setlength{\tabcolsep}{4pt}
\renewcommand{\arraystretch}{1}
\begin{tabular}{lccc cccc}
\toprule
Model 
& RMSE-1 & RMSE-all & \#P 
& MP & T & H & C \\
\midrule
MGN~\cite{Pfaff2020MeshGraphNets}      
& \underline{0.25} & 15.1 & 2.0M 
& \checkmark & $\times$ & \checkmark & \checkmark \\
EvoMesh~\cite{EvoMesh-deng25}          
& 0.28 & 12.9 & 3.2M 
& \checkmark & $\times$ & $\times$ & L \\
HCMT~\cite{ZHOU2025103458}             
& -- & 7.3 & 2.5M 
& $\times$ & \checkmark & $\times$ & L \\
EAGLE~\cite{EAGLE-Janny2023}\cite{M4GN_LeiBo2025}      
& 0.36 & 3.8 & $\approx$ 2.0M* 
& \checkmark & $\times$ & $\times$ & P \\
M4GN~\cite{M4GN_LeiBo2025} 
& 0.26 & \textbf{2.6} & $\approx$ 2.0M* 
& \checkmark & \checkmark & $\times$ & P \\
\midrule
\textbf{MGN-T} (Ours)                  
& \textbf{0.10} & \underline{3.2} & \textbf{0.5M} 
& \checkmark & \checkmark & \checkmark & \checkmark \\
\bottomrule
\end{tabular}
\vspace{2pt}
\caption{Performance comparison on the deforming plate dataset test split. 
($\times 10^{-3}$) Error. \#P: parameters. 
MP: MPNN, T: Transformer, H: hierarchy-free, 
C: coarsening-free (P: prior, L: learnt).  * estimated, values were not reported in the original work.}
\label{tab:plate_combined}
\end{table}

\begin{table}[t]
\centering
\scriptsize
\setlength{\tabcolsep}{6.5pt}
\renewcommand{\arraystretch}{1.}
\begin{tabular}{lcccc}
\toprule
 & \multicolumn{2}{c}{RMSE} & \multicolumn{2}{c}{R-RMSE} \\
\cmidrule(lr){2-3} \cmidrule(lr){4-5}
 & $\bs{q}$ & $\sigma_{\rm VM}$ & $\bs{q}$ & $\sigma_{\rm VM}$ \\
\midrule
All & 3.16$\pm$0.25 & 5.83$\pm$0.34 & 0.63$\pm$0.05 & 3.61$\pm$0.21 \\
1-step & 0.10$\pm$0.005 & 3.87$\pm$0.17 & 0.02$\pm$0.001 & 2.5$\pm$0.13 \\
\bottomrule
\end{tabular}

\vspace{2pt}

\caption{Performance on deforming plate. RMSE ($\times 10^{-3}$) and relative RMSE (\%) for $\bs{q}$ and $\sigma_{\rm VM}$.}
\label{tab:deforming_plate_errors}
\end{table}

\subsection{Pi-Beam}
\label{sec:exp_pibeam}
This benchmark faces two major challenges: the number of degrees of freedom with a 15~k-node mesh and the physical complexity of simulating the entire system for an impact-dynamics problem involving dissipative plastic materials.

Due to the high number of nodes, it is not feasible to satisfy the lower bound on the necessary message passing steps dictated by the criterion established in~\cite{under-reach-mpi}, where a standard MGN would require more than 175 message passing iterations (MPI) to cover the deformable structure. Using fewer iterations leads to \emph{under-reaching}, preventing the network from capturing relevant physics and leading to inaccurate predictions. As a baseline for comparison, we use the original MGN configuration of 15-layer MPNNs (i.e, 15 message-passing iterations) from \cite{Pfaff2020MeshGraphNets}. The \mbox{MGN-T} follows the same architecture as described in section \ref{sec:exp_deforming}, now with $P=64$. Models were trained for 200 epochs, also with a cosine annealing learning rate scheduler.

\begin{figure}[t]
\centering
\begin{subfigure}{\linewidth}
    \centering
    \includegraphics[width=\linewidth]{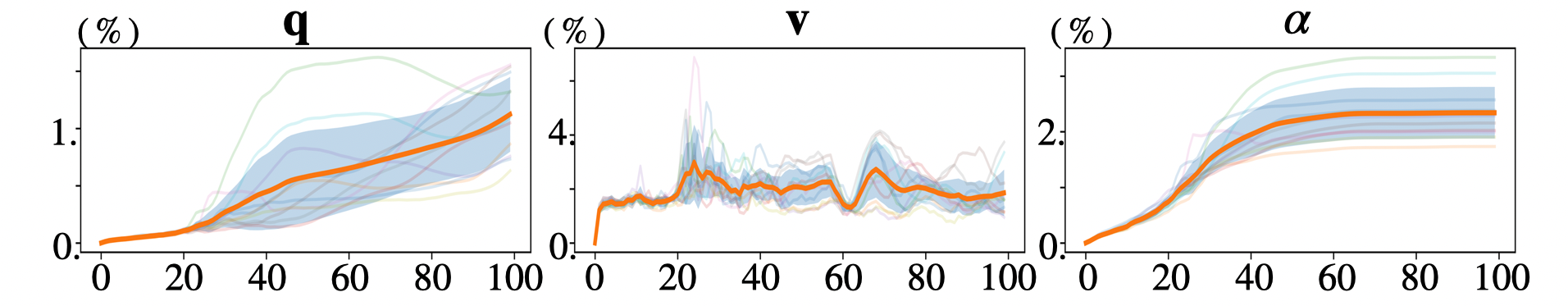}
    \caption{Relative Error evolution}
    \label{fig:rollout_error}
\end{subfigure}

\vspace{0.5em}

\begin{subfigure}{\linewidth}
    \centering
    \includegraphics[width=\linewidth]{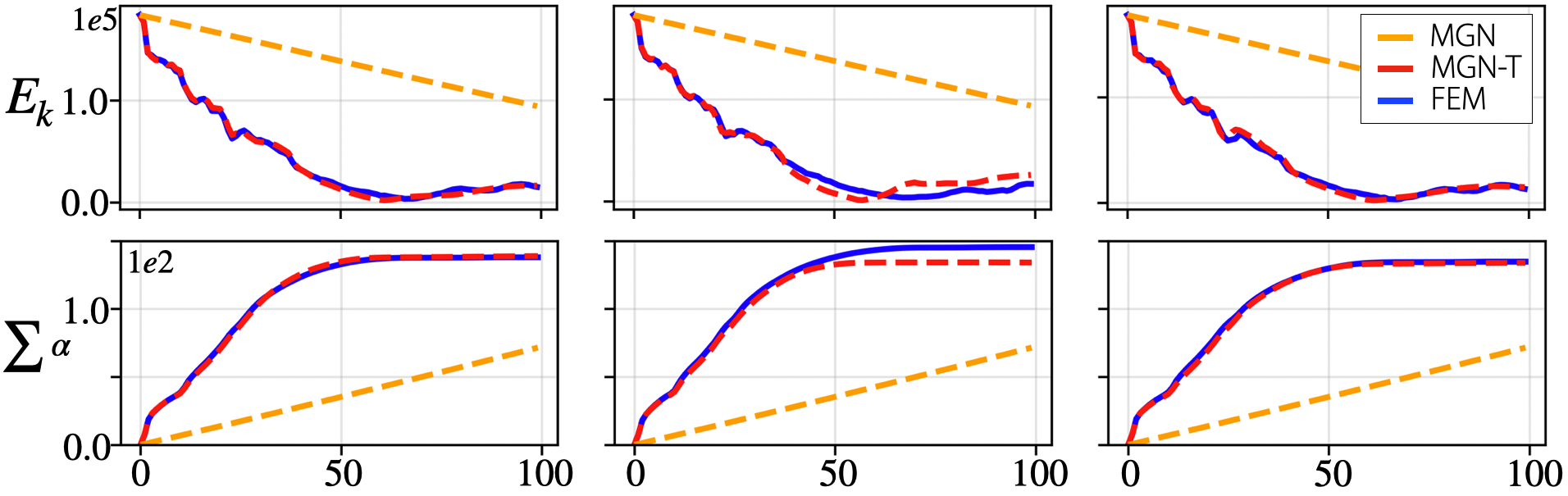}
    \label{fig:physics_pibeam_test}
    \caption{Physical consistency comparison}
\end{subfigure}

\caption{a) Error evolution rollout comparison between FEM ground truth and \mbox{MGN-T} predictions for the Pi-Beam benchmark. Lines represent the mean, and shaded regions the standard error across trajectories. b) Physical consistency comparison MGN, \mbox{MGN-T} vs FEM for three different test trajectories Pi-Beam benchmark.}
\label{fig:rmse_esi_rollouts}

\end{figure}

Table~\ref{tab:unified_all_datasets} reports a comparison of absolute and relative errors. \mbox{MGN-T} outperforms MGN, as it successfully overcomes the under-reaching limitation of MGN, maintaining the relative errors well below 5\%, as shown in Fig.~\ref{fig:rollout_error}. This proves that purely MPNNs, such as MGN, are limited to simulating small-mesh problems far from industrial-scale problems.

\begin{figure}[t]
    \centering
    \begin{subfigure}[t]{0.95\linewidth}
        \centering
        \includegraphics[width=\linewidth]{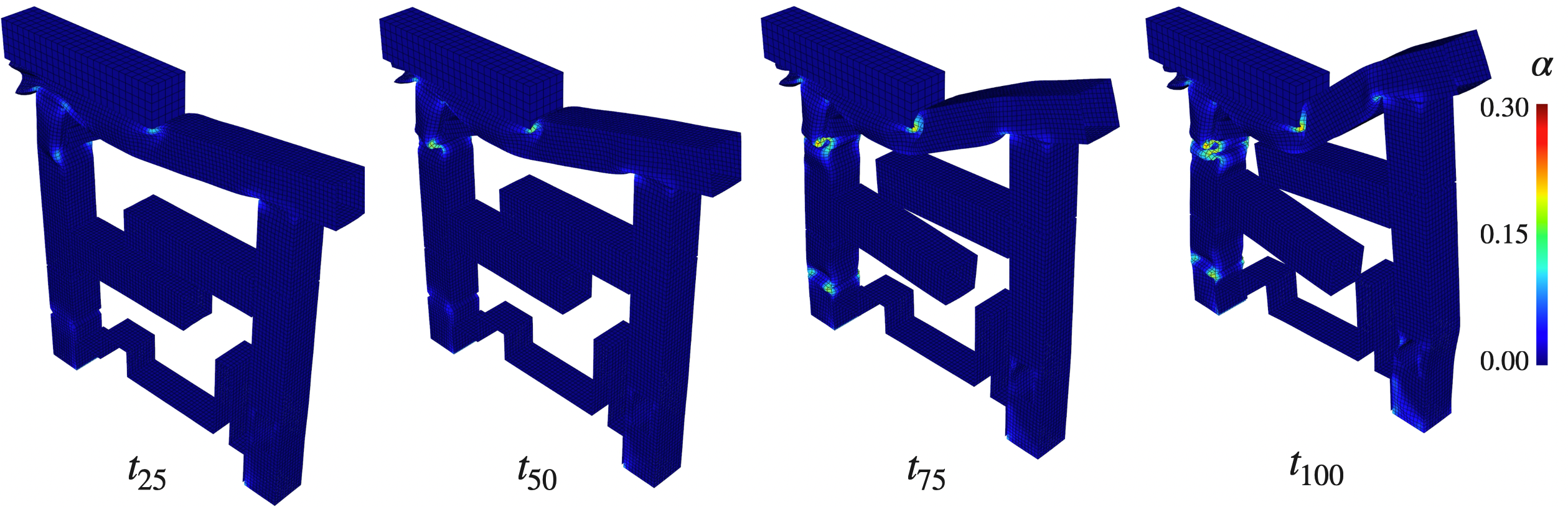}
        \caption{}
        \label{fig:esi_alpha}
    \end{subfigure}
    \vspace{0.3em}

    \begin{subfigure}[t]{0.95\linewidth}
        \centering
        \includegraphics[width=\linewidth]{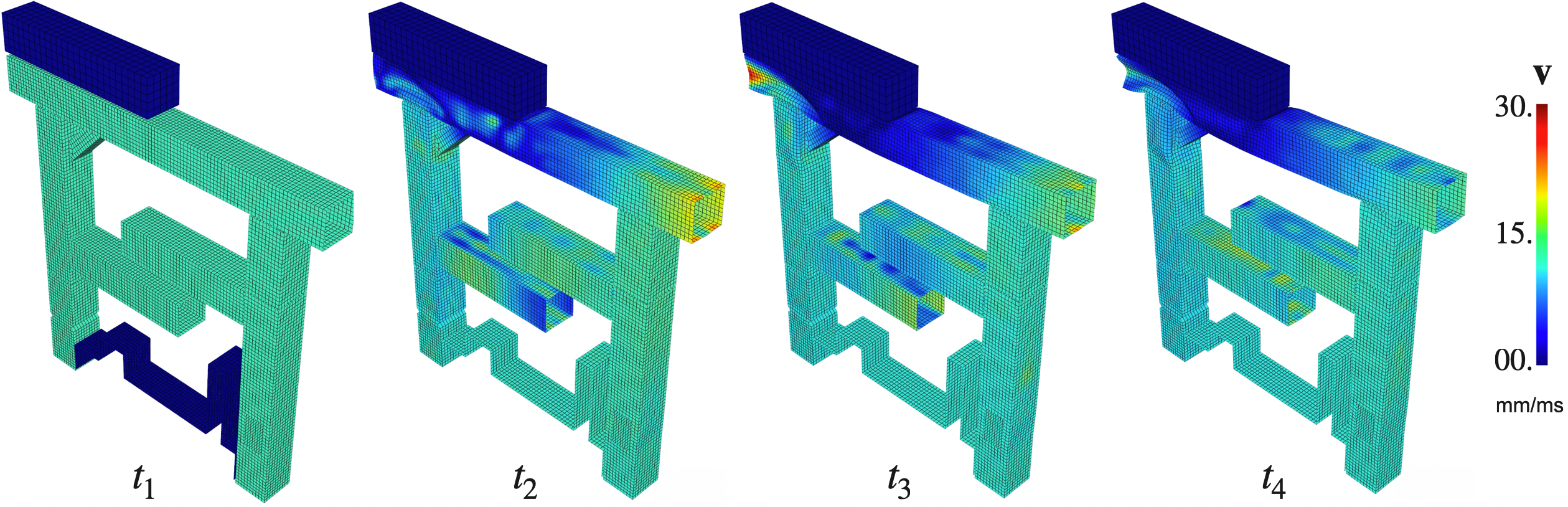}
        \caption{}
        \label{fig:esi_velocity}
    \end{subfigure}

    \caption{Prediction \mbox{MGN-T} Pi-Beam dataset: \mbox{MGN-T} prediction for a dynamic impact against an obstacle. a) Temporal evolution of hardening variable $\alpha$. b) Temporal evolution of velocity magnitude $\|\bs{v}\|$ for consecutive temporal predictions. }
    \label{fig:esi_example}
\end{figure}

Fig.~\ref{fig:esi_alpha} illustrates the prediction of a full trajectory for variable $\alpha$, showing the nodes undergoing plastic deformations. Fig.~\ref{fig:esi_velocity} shows the velocity magnitude evolution at each time step for consecutive predictions. \mbox{MGN-T} accurately predicts buckling, particularly high on the top-left side of the component. These demonstrate the outstanding \mbox{MGN-T} capability to handle nonlinear
dynamics using longer (a $10^4$ factor) temporal steps than FEM. 

\begin{table*}[t]
\centering
\scriptsize
\setlength{\tabcolsep}{7pt}
\renewcommand{\arraystretch}{0.9}

\begin{tabular}{lcccccc}
\toprule
 & \multicolumn{3}{c}{RMSE} & \multicolumn{3}{c}{R-RMSE (\%)} \\
\cmidrule(lr){2-4} \cmidrule(lr){5-7}
Model 
& $\bs{q}$(mm) & $\bs{v}$(mm/ms) & $\alpha$(-) $\times 10^{-3}$
& $\bs{q}$ & $\bs{v}$ & $\alpha$ \\
\midrule

\multicolumn{1}{l}{\textbf{Pi-Beam}} 
\\
\midrule
MGN~\cite{Pfaff2020MeshGraphNets} 
& 86.73$\pm$0.53 
& 4.55$\pm$0.02  
& 23.60$\pm$0.13 
& 6.91$\pm$0.04 
& 14.81$\pm$0.24 
& 7.02$\pm$0.17 \\

\textbf{MGN-T (ours)} 
& \textbf{6.92}$\pm$1.08
& \textbf{0.50}$\pm$0.02  
& \textbf{6.67}$\pm$0.70  
& \textbf{0.55}$\pm$0.08 
& \textbf{1.64}$\pm$0.08 
& \textbf{1.97}$\pm$0.12 \\

\midrule

\multicolumn{1}{l}{\textbf{BIW Crash}} 
& 
&
&
&
& 
& \\
\midrule
MGN-T (P=128)
& 3.96$\pm$0.57  
& 8.02$\pm$0.73  
& 8.42$\pm$0.50
& 0.31$\pm$0.04 
& 2.62$\pm$0.25
& 1.60$\pm$0.11 \\
\bottomrule
\end{tabular}
\caption{
RMSE and relative RMSE (R-RMSE) comparison across datasets. 
Pi-Beam and BIW crash report errors for the physical variables ($\bs{q}, \bs{v}, \alpha$). 
Values are mean $\pm$ standard deviation errors over test trajectories.
}
\label{tab:unified_all_datasets}
\end{table*}

\textbf{Plasticity evolution and physical consistency}: We evaluate physical consistency by analysing the evolution of the plastic hardening variable $\alpha$, which, according to yield criteria, should increase monotonically. As a proxy for measuring the degree of yielding, we visualise the sum of the hardening variable $\alpha$ across nodes at each time step. Similarly, to aggregate the velocity, we calculate the kinetic energy. Fig.~\ref{fig:rmse_esi_rollouts} shows, unlike MGN, how \mbox{MGN-T} closely follows the FEM ground truth solution, preserving monotonic hardening throughout the simulation, in agreement with plasticity theory. The same consistency is observed for the kinetic energy, which decreases as the system dissipates energy during the plastic regime and gathers part of it as potential elastic energy, then is recovered during springback (around time step 50). These results indicate that \mbox{MGN-T} captures key physical constraints of plasticity and the corresponding relationships with other state variables.

\subsection{Body in White crash}
\label{sec:exp_toyota}

\mbox{MGN-T} successfully simulates a large number of nodes in a real-scale industrial crash test, handles multi-material properties, and captures the different dynamics arising from varying thickness setups across trajectories (see Table~\ref{tab:unified_all_datasets}).

\textbf{Ablation study}: We assess the impact of the positional encoding (PE), the number of message passing iterations (MPI) of the pre-processor and refinement blocks and the $P$ tokens, described in Fig.~\ref{fig:sensitivity_ablation}.

\begin{figure}[h!]
\centering
\begin{tikzpicture}
\begin{axis}[
    ybar,
    width=\columnwidth,
    height=5cm,
    bar width=7pt,
    ymin=0,
    ylabel={Error (\%)},
    symbolic x coords={q, v, $\alpha$},
    xtick=data,
    enlarge x limits=0.15,
    ymajorgrids=true,
    tick label style={font=\small},
    label style={font=\small},
    legend style={
        font=\small,
        at={(0.5,1.15)},
        anchor=south,
        legend columns=5
    },
]
\addplot+[] coordinates {(q,2.07) (v,8.39) ($\alpha$,9.8)};
\addplot+[] coordinates {(q,0.95) (v,5.23) ($\alpha$,2.06)};
\addplot+[] coordinates {(q,0.75) (v,4.30) ($\alpha$,2.00)};
\addplot+[] coordinates {(q,0.36) (v,2.75) ($\alpha$,1.72)};
\addplot+[] coordinates {(q,0.415) (v,2.85) ($\alpha$,2.10)};
\legend{No PE, MPI=0, MPI=1, MPI=2, MPI=3}
\end{axis}
\end{tikzpicture}

\vspace{0.3cm}
\begin{tikzpicture}
\begin{axis}[
    ybar,
    width=\columnwidth,
    height=5cm,
    bar width=9pt,
    xlabel={$P$},
    ylabel={Mean RMSE},
    symbolic x coords={0,1,16,32,64,128,256},
    xtick=data,
    ymin=0,
    grid=major,
]
\addplot+[
    fill=gray!35,
    draw=none,
    error bars/.cd,
    y dir=both,
    y explicit,
    error bar style={black}, 
] coordinates {
    (0,13.27) +- (0,1.17)
    (1,11.30) +- (0,1.60)
    (16,5.91) +- (0,0.85)
    (32,4.44) +- (0,0.72)
    (64,4.16) +- (0,0.49)
    (128, 3.99) +- (0,0.57)
    (256,4.05) +- (0,0.6)
};
\end{axis}
\end{tikzpicture}
\caption{
Top: Ablation study. No PE: No Positional Encoder. MPI=0: No MPNN blocks. MPI=1: MPNN with 1 iteration in preprocessor and refinement blocks. MPI=2: with 2 and MPI=3 with 3. Reported errors are shown for each state variable.
Bottom: Ablation study $P$ tokens. RMSE in position for different values of $P$ physical tokens (with PE and MPI=2).}
\label{fig:sensitivity_ablation}

\end{figure}

The positional encoder provides the transformer with explicit spatial information about node locations. This is required to compensate for the loss of graph connectivity during the global processor update, where the physics-attention takes place. The purely mesh-based architecture relies on local relative distances between neighbouring nodes, which are insufficient to provide the nodes with a global spatial structure for feeding the transformer.

On the other hand, the message-passing Pre-processor is responsible for absorbing the boundary conditions. Without pre-processor and refinement blocks (MPI=0), the architecture ignores collisions and self-contact, losing this inductive bias and leading to poor predictions. We observed that more MPIs improve performance, but can lead to overfitting; nevertheless, we believe that more iterations might benefit if more data were available. We found that MPI=2 was the best-performing.

Increasing the number of $P$ tokens decreases the error of the simulation, showing asymptotic behaviour for $P > 32$. Notice that the model performance drops whenever the number of tokens is below this threshold, as it is an excessive reduction from $N$ nodes to $P$ tokens. All experiments were performed under the same training conditions in terms of batch size, epochs and learning rate, employing positional encoder and preprocessor and refinement blocks with MPI=2.

\textbf{Explainable model:} No mathematical theory has yet been developed to determine the optimal number of $P$ tokens. However, moving beyond the black-box perspective of data-driven methods, further analysis can be performed to understand why these techniques work so effectively. 

\begin{figure}[h]
    \centering
    \includegraphics[width=\linewidth]{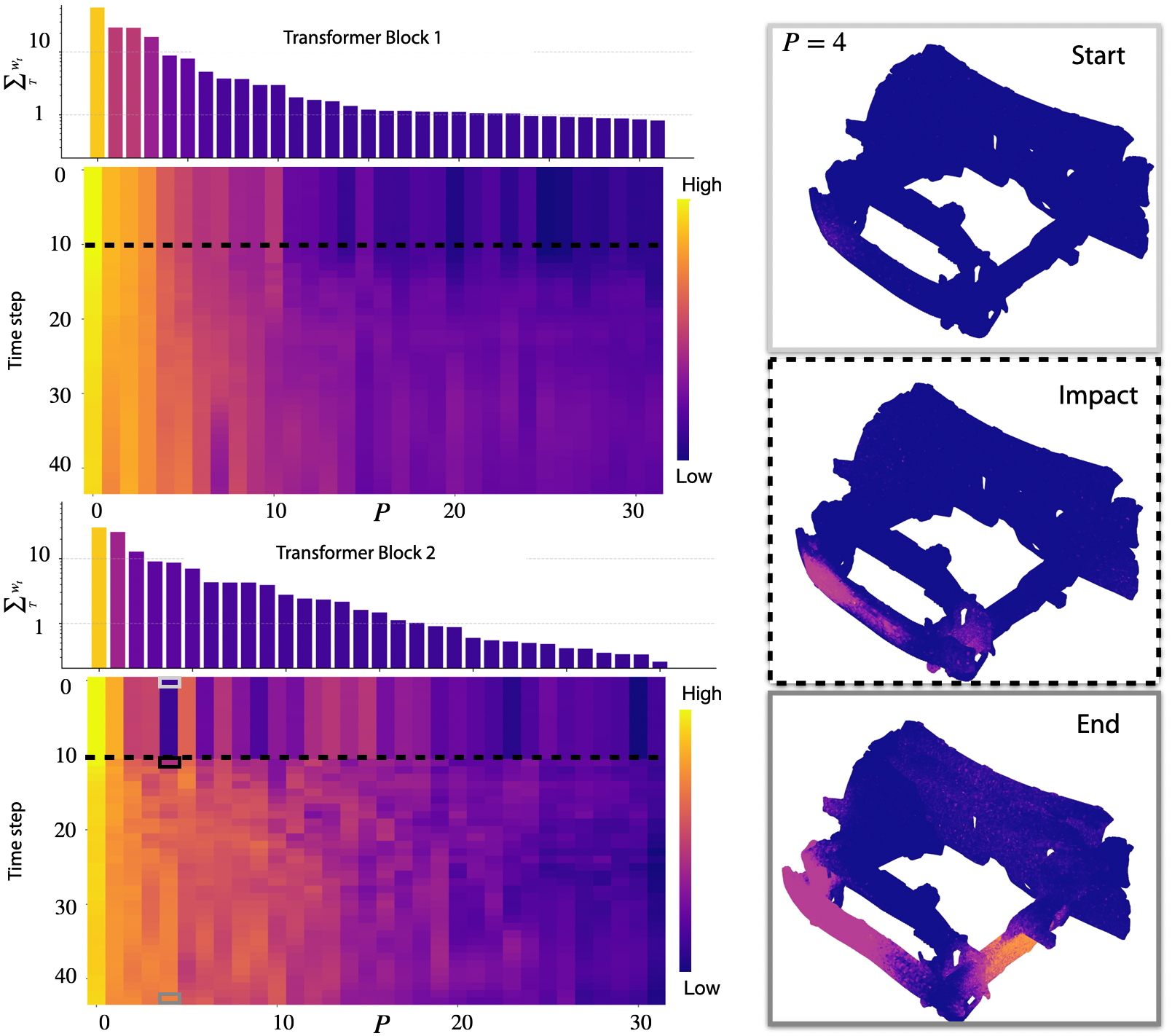}
   \caption{Left: Transformer blocks 1 and 2. Each heatmap shows the temporal evolution of the weights, where each cell represents the total sum of weights $w_i$ per node associated with each token ($x$-axis) at a given time step ($y$-axis). Above each heatmap, the weight contributions to each token $P$ are calculated as the sum across all time steps. Tokens are ordered from highest to lowest contribution. The dashed line marks the moment of impact. Right: Evolution of the $P=4$ weights along the simulation.}
    \label{fig:tokens_importance}
\end{figure}

Fig.~\ref{fig:tokens_importance} illustrates the learned dynamic hierarchical representation of the P tokens, where the weights adapt to positional information and physical state variables, as reflected by the variation along the y-axis, showing their temporal evolution. 

A closer observation of Fig.~\ref{fig:tokens_importance} reveals two phenomena. First, the transformer blocks have different behaviour: the first tends to have smoother changes over time, while the second tends to have sharper changes. This may be a consequence of the first block focusing on steadier phenomena, such as geometry, while the second block focuses on abrupt changes, such as velocity $\bs{v}$ or impact. In fact, it can be observed that in the second transformer block heatmap, token weights remain constant until the vehicle impacts the rigid wall-- dashed line at time step=10-- the moment at which the heatmap changes abruptly, activating and deactivating different weights.
Second, when analysing the weight contribution to each token, these remain higher in the first block, while they decay more in the second block.

\subsection{Extrapolation: Moving wall}
\label{sec:exp_extrapol}

\begin{figure*}[h]
    \centering
    \includegraphics[width=\linewidth]{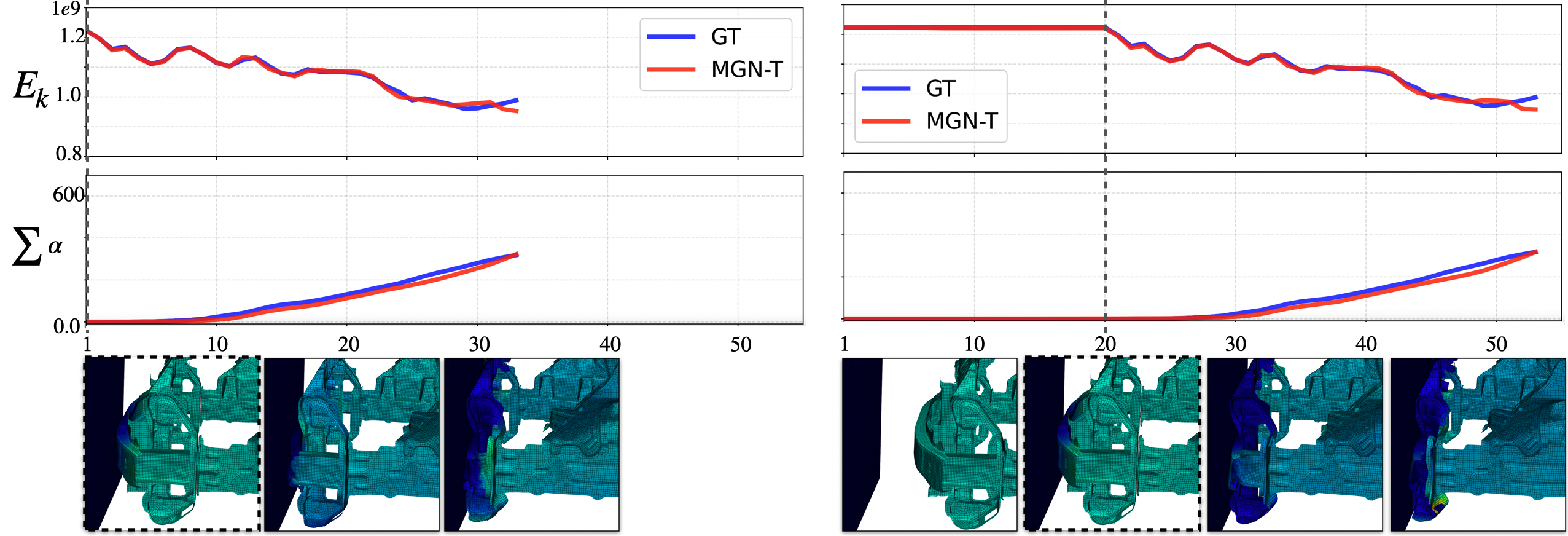}
   \caption{Extrapolation moving wall test example. Kinetic energy ($E_k$) and accumulated plastic hardening ($\sum \alpha$) evolution over time. Left: Wall located 1 step from the BIW. Right: wall located 20 steps from the BIW. Dashed lines represent the moment of impact. Below the plot, a visualisation of predictions in colour of the velocity magnitude $||\bs{v}||$.}
\label{fig:extrapolation}
\end{figure*}

It is well-known that data-driven methods tend to struggle with extrapolation; therefore, in practice, the key to achieving good results in such scenarios is to keep the model within the interpolation regime. The \textit{moving wall} setup is an extrapolation experiment where the location of the wall changes from the location used during training, fixed at 10 time steps from the start. This experiment demonstrates that our method has learnt to detect impact and simulate the resulting dynamics independently of the starting location. As shown in Fig.~\ref{fig:extrapolation}, dynamics after impact closely match across different location cases. Moreover, despite of being of an autoregressive character---where error accumulation affects the quality of the predictions---, the kinetic energy $E_k$ and plastic hardening remain stable prior to impact, proving its physical consistency in this out-of-distribution example.

Thanks to the autoregressive mesh-based architecture of \mbox{MGN-T}, the multi-graph $G(V, E_M, E_C)$ varies dynamically, as contact edges are recomputed after each roll-out iteration. The subgraphs---rigid wall and car---will only become connected through the contact edges if relative distances between nodes are sufficiently short. This naturally brings the model to remain within the interpolation regime when performing extrapolation tasks (i.e., the model never really knows where the car and wall are located in space, just whether they are in contact or not, as no explicit locations are required).

\subsection{Computational Cost}

All experiments were executed on a single NVIDIA RTX 4090 GPU (24 GB) of memory. Inference time consumption is described in Table~\ref{tab:computational_cost}. From a hardware perspective, despite the ability to simulate large meshes, GPU memory may not meet the computational demands of much larger meshes.

\begin{table}[t] 
\centering
\small
\setlength{\tabcolsep}{4pt}
\renewcommand{\arraystretch}{0.9}
\begin{tabular}{@{}lrrr@{}}
\toprule
Dataset \& Model & \makecell{$t_{\text{model}}$ \\ (ms/step)} & \makecell{$t_{\text{full}}$ \\ (ms/step)} & \makecell{Memory \\ (MB)} \\
\midrule
Deforming Plate MGN-T & 2.3 & 3.2  & 1\,752  \\
Pi-Beam MGN          & 14.0   & 35.0   & 12\,512 \\
Pi-Beam MGN-T        & 4.1  & 15.0   & 1\,782  \\
BIW MGN-T        & 8.9  & 55.0   & 2\,315  \\
\bottomrule
\end{tabular}
\caption{Computational cost and memory employed in inference.}
\label{tab:computational_cost}
\end{table}

\section{Discussion}\label{discussion}

In our experiments, \mbox{MGN-T} successfully simulates industrial-scale impact dynamics on meshes with up to 100~k nodes, comprising more than 100 components and 17 different materials, including dissipative plastic behaviour. To the best of our knowledge, we are the first to simulate such scenarios with a mesh-based graph architecture. We demonstrate the benefits of keeping the mesh-based representations through the Message Passing Neural Network (MPNN) blocks, and using a transformer with physics attention as a global update processor to overcome the limitation of message-passing under-reach. 

We further analyse the impact of different hyperparameters, $P$ tokens, message-passing iterations, and the positional encoder, providing insights into what the model has learnt by analysing the latent $P$ token weights $w$ and their contributions. { \mbox{MGN-T} combines the strengths of MPNNs and transformers: the MPNN component effectively captures local interactions with little MPI, such as contact phenomena, while the transformer component provides the scalability required to handle large meshes.}

In addition, we showed that \mbox{MGN-T} achieves competitive results compared to state-of-the-art approaches, demonstrating low error propagation in both single-step and full-rollout predictions on the Deforming Plate benchmark, ranking second-best. Compared to other approaches, \mbox{MGN-T} requires a fraction (roughly 1/4) of the parameters, making it a much lighter and faster model. 

\mbox{MGN-T} addresses a fundamental bottleneck in MPNNs and enables training on large, complex meshes without requiring mesh coarsening or a hierarchy. This facilitates training on industrial-scale datasets for real-world engineering applications, enabling the utilisation of simulated data with minimal human intervention. We can conclude that \mbox{MGN-T} stands out as a powerful architecture for simulating solid mechanics, merging the best of mesh-based and transformer methods.

\subsection{Limitations and Future Work}


This work adopts a purely data-driven learning paradigm and does not incorporate explicit physical constraints during training, lacking physical guarantees at inference time. Future research directions focus on tackling this major limitation.
Second, the model’s ability to generalise to unseen thickness configurations depends on the diversity and coverage of the training data. Although the proposed method shows promising extrapolation capabilities in the moving-wall experiments, its performance on thickness values, geometries or wall locations that lie outside the training distribution cannot be guaranteed.
Finally, the current contact feature engineering is based on Euclidean distances between mesh nodes, which may include physically impossible interactions, such as contacts through structures. Restricting these to physically feasible interactions would reduce the graph size and improve computational efficiency. However, it remains unclear whether enforcing only physically admissible contacts would improve predictive performance, as models appear to benefit from the additional contextual information.

\section*{Acknowledgements}
This work was supported by the Spanish Ministry of Science and Innovation, AEI/10.13039/501100011033, through Grant number PID2023-147373OB-I00, and by the Ministry for Digital Transformation and the Civil Service, through the ENIA 2022 Chairs for the creation of university-industry chairs in AI, through Grant TSI-100930-2023-1.

Funded by the European Union. Views and opinions expressed are however those of the author(s) only and do not necessarily reflect those of the European Union or the European Research Council Executive Agency. Neither the European Union nor the granting authority can be held responsible for them. This
work is supported by ERC grant PHYSIA 101264273.

The authors also acknowledge the support of ESI Group, Keysight Technologies, through the chair at the
University of Zaragoza.

\appendix
\section*{Appendix}

\section{Errors}
\label{ap:errors}
RMSE is computed as the root-mean-square error between predictions and FEM solutions, taken as ground truth, per variable. For each error, we calculate the mean squared error across all feature variables, nodes, time steps, and trajectories, then take the square root (Described in \cite{Pfaff2020MeshGraphNets}, Appendix A.5.2 ). RMSE-1 is computed without autoregressive rollout: predictions always start from the ground-truth state to prevent error accumulation. Relative RMSE (R-RMSE) follows the same computation, with each error normalized by the infinity norm of the corresponding trajectory (i.e., its maximum magnitude). 

\end{document}